\pdfoutput=1

\documentclass[11pt]{article}

\usepackage[preprint]{acl}

\usepackage{times}
\usepackage{latexsym}

\usepackage[T1]{fontenc}

\usepackage[utf8]{inputenc}

\usepackage{microtype}

\usepackage{inconsolata}

\usepackage{graphicx}
\usepackage{hyperref}
\usepackage{url}
\usepackage{booktabs}
\usepackage[flushleft]{threeparttable}
\usepackage{subcaption}
\usepackage{colortbl}
\usepackage{makecell}
\usepackage{multirow} 

\usepackage{amsmath} 

\title{Towards Better Evaluation for Generated Patent Claims}

\author{
Lekang Jiang$^{\dagger}$,  
Pascal A Scherz$^{\diamond}$,
Stephan Goetz$^{\dagger}$ \\
$^{\dagger}$University of Cambridge, $^{\diamond}$PSPB Patent Law\\ 
\texttt{\{lj408, smg84\}@cam.ac.uk, post@pspb.eu}
}

\begin{document}
\maketitle
\begin{abstract}

Patent claims define the scope of protection and establish the legal boundaries of an invention. Drafting these claims is a complex and time-consuming process that usually requires the expertise of skilled patent attorneys, which can form a large access barrier for many small enterprises. To solve these challenges, researchers have investigated the use of large language models (LLMs) for automating patent claim generation. However, existing studies highlight inconsistencies between automated evaluation metrics and human expert assessments.
To bridge this gap, we introduce Patent-CE, the first comprehensive benchmark for evaluating patent claims. Patent-CE includes comparative claim evaluations annotated by patent experts, focusing on five key criteria: feature completeness, conceptual clarity, terminology consistency, logical linkage, and overall quality. Additionally, we propose PatClaimEval, a novel multi-dimensional evaluation method specifically designed for patent claims. Our experiments demonstrate that PatClaimEval achieves the highest correlation with human expert evaluations across all assessment criteria among all tested metrics. This research provides the groundwork for more accurate evaluations of automated patent claim generation systems.\footnote{\url{https://github.com/scylj1/PatClaimEval}}

\end{abstract}

\section{Introduction}

\begin{table*}[!ht]
\centering
\footnotesize
\resizebox{\textwidth}{!}{

\begin{tabular}{|p{15.5cm}|}
\toprule
\textbf{Reference claims:} \\
1. A mobile communications device comprising: \\
a communication subsystem \textcolor{blue}{for} communicating with a network; \\
a \textcolor{blue}{microprocessor} operably connected to the communication subsystem and to a memory; \\
\textcolor{blue}{a local common address database accessible to a plurality of applications on the mobile communications device; and} \dots \\
2. The mobile communications device of claim 1 wherein  \dots \\ 
\dots \\
\textbf{Candidate claims:} \\
1. A mobile communications device comprising: \\
a communication subsystem \textcolor{blue}{configured to facilitate} communication with a network; \\
a \textcolor{blue}{processor} operably connected to the communication subsystem and to a memory, \\
\textcolor{blue}{the memory containing a local common address database and instructions that} \dots \\
2. The mobile communications device of claim 1, wherein \dots \\
\dots \\
\midrule

\textbf{N-gram-based evaluations (measuring sequence overlaps):} \\
BLEU \quad ROUGE \quad METEOR \quad  \dots \\
\midrule

\textbf{Embedding-based evaluations (measuring semantic similarities):} \\
BERTScore \quad BARTScore \quad MoverScore \quad SimCSE \quad  \dots \\
\midrule

\textbf{Multi-dimensional evaluations:} \\
UniEval (Coherence, Consistency, Fluency, Relevance) \quad AlignScore (Factual consistency) \quad  \dots \\
\midrule

\textbf{Human evaluations for patent claims:} \\
Feature completeness \quad Conceptual clarity \quad Terminology consistency \quad Logical linkage \quad Overall quality \\

\bottomrule
\end{tabular}
}
\caption{Comparison between current automatic text evaluation metrics and patent claim evaluation criteria. Patent claims have specific requirements different from other texts, which makes the evaluation difficult.  }
\label{table:com}
\end{table*}

The patent literature serves as a critical documentation of technological innovation \citep{mossoff2000rethinking}. Patents are legal documents that grant exclusive rights to inventors in exchange for public disclosure of their inventions \citep{Frumkin1947}. The patent claims are the most legally significant section of a patent document, as they define the scope of protection and delineate the boundaries of an invention from known techniques to ensure legal enforceability \citep{epo2020}. Drafting precise and effective patent claims is a challenging task, which requires not only technical expertise but also an understanding of legal language and jurisdiction-specific regulations \citep{faber1990landis}. Unlike general-purpose texts, patent claims must be both broad enough to encompass potential variations of an invention and specific enough to withstand legal scrutiny. This complexity often necessitates the involvement of skilled patent attorneys, which often renders the process both time-intensive and costly \citep{CisloAndThomas2023}.

In response to these challenges, research has explored automated methods for patent claim generation to support inventors and attorneys. Large language models (LLMs) have demonstrated remarkable capabilities across a wide range of tasks \citep{zhao2023survey}, including tasks in the patent domain \citep{jiang2024artificial}. For instance, \citet{jiang-etal-2025-large} examined whether LLMs could generate high-quality patent claims based on patent descriptions, while another work investigated whether LLMs could revise patent claims to improve quality \citep{jiang-etal-2025-patent}. These studies aim to accelerate the claim drafting process and reduce associated costs.

Despite these advances, the reliable automated evaluation of the quality of generated patent claims remains an unresolved challenge. Previous studies have relied on human expert evaluations, which are both time-consuming and costly, and they revealed inconsistencies between existing automated metrics and human assessments \citep{zuo-etal-2024-patenteval, jiang-etal-2025-large, jiang-etal-2025-patent}. Table \ref{table:com} highlights the fundamental differences between patent claim evaluation criteria and existing text evaluation methods. While current evaluations often rely on sequence overlap, semantic similarity, or multi-dimensional criteria such as coherence and fluency, patent claims have unique language requirements. Such requirements, such as the consistent use of terminology, technical formality, and high-level precision, are not common in other types of texts. Therefore, these differences underscore the limitations of existing metrics and highlight a significant gap in the reliability and validity of automated evaluation methods for patent claims.

In this paper, we present the first benchmark for patent claim evaluation and propose a novel evaluation method tailored to the unique requirements of patent claims.
The main contributions of this work are as follows:

\noindent 1. We present Patent-CE, the first comprehensive benchmark for patent claim evaluation. Patent-CE includes 1,228 data points, which consist of a reference claim, two candidate claims, and comparative evaluations annotated by patent experts based on feature completeness, conceptual clarity, terminology consistency, logical linkage, and overall quality.

\noindent 2. We propose a novel multi-dimensional evaluation method for patent claims, named PatClaimEval. PatClaimEval leverages Longformer \citep{beltagy2020longformer} as its backbone and is trained on our dataset using a variation of contrastive learning \citep{gao-etal-2021-simcse}.

\noindent 3. We demonstrate the effectiveness of PatClaimEval through extensive experiments. Our results show that PatClaimEval achieves the highest correlation with human expert evaluations across all assessment criteria compared to existing metrics, including six N-gram-based methods, four embedding-based approaches, two multi-dimensional evaluators, and one LLM-as-a-judge method.

By tackling the evaluation problem, this research paves the way for more reliable assessments of automated patent claim generation systems, ultimately contributing to advancements in this emerging field.\footnote{Notably, we focus on reference-based claim evaluations, which is different from the real patent examination. We introduce details in the Limitation section. }

\section{Related Works}

\begin{table*}[!ht]
\centering
\footnotesize
\begin{tabular}{l|c|c|p{4.8cm}}
\toprule
\textbf{Dataset} & \textbf{Task} &\textbf{Domain} & \multicolumn{1}{c}{\textbf{Evaluation Criteria}} \\
\midrule
QAGS \citep{wang-etal-2020-asking} & Summarization & News & Factual consistency \\ \midrule

\multirow{2}{*}{SummEval \citep{fabbri-etal-2021-summeval}} & \multirow{2}{*}{Summarization} & \multirow{2}{*}{News} & Fluency, coherence, consistency, relevance \\ \midrule

\multirow{2}{*}{Persona-Chat \citep{zhang-etal-2018-personalizing}} & \multirow{2}{*}{Dialogue generation} & \multirow{2}{*}{General} &  Fluency, engagingness, consistency, personalization \\ \midrule

\multirow{2}{*}{Topical-Chat \citep{mehri-eskenazi-2020-usr}} & \multirow{2}{*}{Dialogue generation} & \multirow{2}{*}{General} &  Naturalness, coherence, engagingness, groundedness, understandability \\ \midrule 

ToTTo \citep{parikh-etal-2020-totto} & Table-to-text generation & General & Fluency, faithfulness, coverage \\ \midrule 

\rowcolor{gray!30}
\multirow{3}{*}{Patent-CE (Ours)} & \multirow{3}{*}{Patent claim generation} & \multirow{3}{*}{Patent} & Feature completeness, conceptual clarity, terminology consistency, logical linkage, overall quality  \\
\bottomrule
\end{tabular}
\caption{Comparison of commonly used benchmarks for text generation evaluation. Patent claims have unique language requirements different from other type of texts. }
\label{table:datasets}
\end{table*}

\subsection{Patent Claim Generation}

Some studies have explored the use of LLMs for automatically generating patent claims. An early investigation by \citet{lee2020patentgenerate} served as a preliminary effort and focused on fine-tuning GPT-2 \citep{radford2019language} to generate patent-like texts. The authors found that minimal training was sufficient to produce patent-like outputs but did not assess the quality of the generated text. Building on this, \citet{lee2020controlling} trained GPT-2 to transform one section of a patent application into another, such as the generation of abstracts from titles or claims from abstracts. However, since abstracts are often generic and imprecise, the generation of claims from abstracts may not be a well-conditioned task.

Therefore, \citet{jiang-etal-2025-large} introduced a description-based claim generation task and evaluated the performance of various LLMs on this domain-specific challenge. Their human evaluation by patent professionals highlighted the limitations of various LLMs in generating high-quality patent claims and revealed inconsistencies between automated and human evaluation metrics. Additionally, \citet{jiang-etal-2025-patent} extended the research to claim revision, investigating whether LLMs could further enhance claim quality.

Another study proposed the task of next-claim generation \citep{zuo-etal-2024-patenteval}, which involves generating the second and/or third claims based on the first claim. Likewise, this research also demonstrated a weak correlation between automated and human evaluation results for the next-claim generation task.

\subsection{Benchmarks for Text Generation Evaluation}

Accurately and efficiently evaluating the quality of generated texts is important for developing text-generation LLMs. Researchers have introduced several text evaluation datasets across various domains, each with distinct evaluation criteria. Table~\ref{table:datasets} summarizes some commonly used benchmarks for text generation evaluations. 

Datasets, such as QAGS \citep{wang-etal-2020-asking} and SummEval \citep{fabbri-etal-2021-summeval}, are designed to evaluate summarization tasks within the news domain. These benchmarks focus on criteria including factual consistency, fluency, coherence, and relevance to ensure the generated summaries accurately represent the source text while maintaining readability.
In addition, dialogue systems have benefited from benchmarks such as Persona-Chat \citep{zhang-etal-2018-personalizing} and Topical-Chat \citep{mehri-eskenazi-2020-usr}, which target open-domain conversational tasks. Persona-Chat emphasizes personalization, fluency, and engagingness, while Topical-Chat introduces evaluation metrics for naturalness, coherence, and groundedness to advance the development of more realistic and context-aware conversational AI.
Furthermore, the ToTTo dataset \citep{parikh-etal-2020-totto} supports the task of converting structured tables into natural text. It evaluates fluency, faithfulness, and coverage to ensure the generated text aligns accurately with the tabular input and effectively conveys the intended information.

Our Patent-CE dataset is specifically designed for the task of patent claim generation. Unlike other benchmarks, Patent-CE emphasizes feature completeness, clarity, terminology consistency, and logical linkage. All these critical aspects are for the legal and technical precision required in patent documentation. This dataset fills an essential gap by providing a benchmark tailored to the patent domain, presenting unique challenges not encountered in general-domain tasks.

\section{Dataset}

\subsection{Human Annotation}
Experienced patent experts were provided with reference and candidate patent claims to evaluate. Their evaluation was based on five aspects, adhering to established examination criteria \citep{jiang-etal-2025-large}: feature completeness, conceptual clarity, terminology consistency, logical linkage, and overall quality. 
These evaluation aspects are consistent with previous research and defined as follows.

\textbf{(1) Feature Completeness}: The extent to which the generated claims encapsulate all critical aspects of the invention.
\textbf{(2) Conceptual Clarity}: The clarity and unambiguity of the language used in the claims.
\textbf{(3) Terminology Consistency}: The uniformity in the use of terms throughout the claims.
\textbf{(4) Correctness of Logical Linkages}: The accuracy with which features are interconnected and related.
\textbf{(5) Overall Quality}: An aggregate measure that combines all the above criteria.
Detailed evaluation instructions can be found in Appendix \ref{humanannotation}.

\subsection{Construction}

To create a comprehensive dataset and mitigate potential biases, we collected data from three different sources.

First, we used the dataset from \citet{jiang-etal-2025-large}, in which LLMs were used to generate patent claims based on descriptions from the United States Patent and Trademark Office (USPTO). This dataset also includes human evaluations which compare the performance of different models. Second, we incorporated data from another study that investigated patent claim revision using data from the European Patent Office (EPO) and also included human evaluations \citep{jiang-etal-2025-patent}. Both studies rated claims based on feature completeness, conceptual clarity, terminology consistency, logical linkage, and overall quality. We integrated these data to construct a comprehensive evaluation benchmark. Additionally, to further increase the dataset size and enhance robustness, we conducted new annotations by consulting patent attorneys. These additional annotations were applied to claims obtained from the aforementioned studies.

We recognize that the absolute quality scores from different sources may vary due to differences in expert interpretation. However, the relative ranking of the same claim sets should remain rather consistent across evaluations. Therefore, similar to prior work by \citet{zuo-etal-2024-patenteval}, our dataset focuses on comparative evaluations. Each data point consists of a quadruplet $(A, B, C, y)$, where $A$ represents the reference claims, $B$ and $C$ are two generated claims, and the label $y$ indicates whether $B$ or $C$ is better, or if they are of equal quality.

\subsection{Statistics}

The dataset consists of a total of 1,228 data points evaluated in five aspects. As shown in Table~\ref{tab:data_stats}, the data distribution is relatively balanced for each aspect, with similar proportions in each category. Appendix~\ref{moredatastats} introduces more dataset statistics. We randomly selected 184 examples (about 15\%) as the test set and used the remaining for training. 

Our benchmark offers two major advantages: \textbf{Comprehensiveness}: It incorporates patent data from multiple patent offices, which makes it more representative and robust than any previous work. \textbf{Larger scale}: Although some data builds on previous work, we manually refine and annotate more data to substantially expand the dataset size and broaden coverage.

\begin{table}[!t]
\centering
\footnotesize

\begin{tabular}{lcccc}
\toprule
\textbf{Dimension} & \textbf{\# (B > C)} & \textbf{\# (B = C)} & \textbf{\# (B < C)} \\
\midrule
Completeness & 426  & 375 & 427  \\
Clarity & 424  & 378 & 426   \\
Consistency & 420  & 386 & 422   \\
Linkage & 430  & 366 & 432    \\
Quality & 422  & 382 & 424   \\
\bottomrule
\end{tabular}

\caption{Data distribution of the Patent-CE dataset. The data distribution is relatively balanced in each evaluation dimension.}
\label{tab:data_stats}
\end{table}

\section{Method}

We propose PatClaimEval, a new automated evaluation method to assess the quality of generated patent claims compared to gold claims. Given a reference claim set $P$ and a candidate claim set $Q$, the model predicts a quality score of $Q$, denoted as $s(Q|P)$. We train five models to evaluate patent claims from different aspects, including feature completeness, conceptual clarity, terminology consistency, logical linkage, and overall quality. We do not jointly train one model for all five aspects, such as using multi-task learning \citep{zhang2021survey}, because of the conflicting optimization objectives for different tasks. For example, feature completeness and clarity are not inherently related---the claim could include all essential features, but the expression is ambiguous.

\subsection{Model Architecture}
We leverage Longformer \citep{beltagy2020longformer} as the backbone to handle long input sequences efficiently. We have not used patent-specific LLMs due to their limitations; for instance, PatentGPT is closed-source \citep{bai2024patentgpt}, and PatentGPT-J has a restricted context length \citep{lee2023evaluating}. The small context length is a particular problem for patent texts, as it may fall short of typical patent claims. The average length of patent claims is more than 1,000 tokens \citep{suzgun2023harvard}, and Longformer can support up to 4,096 tokens of input. Thus, we use Longformer because it is open source, supports long input length (large enough for patent claims), and offers a controllable model size. The model encodes inputs and obtains the representation for a given input claim pair $(P, Q)$ per
\begin{equation}
    \mathbf{h} = \mathcal{M}([P; Q]).
\end{equation}
It subsequently connects with a fully connected layer to get a quality score and a sigmoid function that maps the score to the range $[0, 1]$ as
\begin{equation}
    s(Q|P) = \sigma(\mathbf{w}^\top \mathbf{h} + b).
\end{equation}

\subsection{Training Strategy}

Our proposed training method draws inspiration from contrastive learning \citep{khosla2020supervised} as the dataset presents relative relationships between samples.
In the context of NLP, contrastive learning is used to align embeddings of related text pairs or to learn discriminative representations \citep{gao-etal-2021-simcse}. Through contrastive loss functions, models can capture nuanced differences between text samples, making contrastive learning particularly suitable for tasks involving relative comparisons. Unlike traditional contrastive learning, which explicitly constructs positive and negative sample pairs, our method integrates label information directly to define optimization objectives tailored for different evaluation aspects.

The training data consists of quadruplets $(A, B, C, y)$, where $A$ is the reference claims, $B$ and $C$ are two generated claims, and $y \in \{1, 0, -1\}$ indicates their relative quality:
\begin{equation}
    y = 
    \begin{cases} 
        1, & \text{if } s(B|A) > s(C|A) \\
        0, & \text{if } s(B|A) = s(C|A) \\
        -1, & \text{if } s(B|A) < s(C|A)
    \end{cases}
\end{equation}

The model computes scores $s_B = s(B|A)$ and $s_C = s(C|A)$. To optimize the model, we define the loss function as
\begin{equation}
    \mathcal{L} = \frac{1}{N} \sum_{i=1}^N \ell(s_{B_i}, s_{C_i}, y_i),
\end{equation}
where $\ell(s_{B_i}, s_{C_i}, y_i)$ is defined as
\begin{equation}
    \ell = 
    \begin{cases}
        ReLU(m - (s_{B_i} - s_{C_i})), & \text{if } y_i = 1, \\
        ReLU(|s_{B_i} - s_{C_i}| - n), & \text{if } y_i = 0, \\
        ReLU \left(m - \left(s_{C_i} - s_{B_i}\right)\right), & \text{if } y_i = -1, \\
    \end{cases}
\end{equation}
where $m$ is the margin hyper-parameter that enforces a minimum separation between scores for distinct quality levels, and $n$ is a tolerance parameter that allows small differences between scores when the two claim sets are judged equally good.

By minimizing this loss, the model learns to align the predicted scores with the relative quality judgments. The margin $m$ is a hyper-parameter that controls the separation between scores, ensuring that the model is confident in its predictions for cases where one claim set is clearly better than the other.
This objective function allows the model to capture fine-grained distinctions in quality across diverse claim pairs. We introduce the training and evaluation details in Appendix \ref{experimentdetails}.

\section{Experiments}
\subsection{Baselines}

\textbf{BLEU} \citep{papineni-etal-2002-bleu} and \textbf{ROUGE} \citep{lin2004rouge} are classic metrics widely used for evaluating text overlaps. BLEU measures n-gram precision by comparing candidate and reference texts, while ROUGE primarily evaluates recall-based overlap, commonly used in summarization.
\textbf{METEOR} \citep{banerjee-lavie-2005-meteor} improves on BLEU by incorporating synonymy, stemming, and other linguistic features, thereby providing a more flexible approach to measuring textual overlap.

\textbf{BERTScore} \citep{zhang2019bertscore} computes similarity using contextualized embeddings from BERT \citep{devlin-etal-2019-bert}, enabling a more nuanced assessment of semantic similarity between reference and candidate sentences. \textbf{BARTScore} \citep{NEURIPS2021_e4d2b6e6}, derived from the BART model \citep{lewis-etal-2020-bart}, uses a generative scoring approach to evaluate the likelihood of generating the candidate text from a reference. \textbf{MoverScore} \citep{zhao-etal-2019-moverscore} measures the semantic similarity by calculating the minimum cost of transforming candidate embeddings to reference embeddings, effectively capturing semantic alignment. \textbf{SimCSE} \citep{gao-etal-2021-simcse} further enhances representation quality by using contrastive learning to generate sentence embeddings, which have been shown to perform well in semantic similarity tasks.

We also test recent multi-dimensional evaluation frameworks, including \textbf{UniEval} \citep{zhong-etal-2022-towards} and \textbf{AlignScore} \citep{zha-etal-2023-alignscore}. UniEval provides a unified evaluation protocol for different aspects of natural language generation, such as coherence and fluency. Since our dataset does not include context information or source texts, we use UniEval to evaluate the relevance between generated and reference claims. Additionally, we use AlignScore as a representative to assess factual consistency between source and generated content. 

The LLM-as-a-judge paradigm is becoming popular \citep{zheng2023judging}, where LLMs are used as evaluators of generated content. This approach leverages the capabilities of pre-trained LLMs, such as GPT-4 \citep{achiam2023gpt}, to serve as surrogate judges that can assess generated text for qualities like fluency, coherence, and factual consistency. In our experiments, we specifically focus on \textbf{G-Eval-4} \citep{liu-etal-2023-g} because it has shown high agreement with human preference across multiple benchmarks \citep{zheng2023judging}. Other LLM-as-a-judge models are not tested because they use synthetic examples generated by GPT-4 for training, such as JudgeLM \citep{zhu2023judgelm} PandaLM \citep{pandalm2024}. We ask GPT-4 to evaluate the given claims through chain-of-thought (CoT) prompting \citep{wei2022chain} by comparing them to the reference claims. The evaluation dimensions are the same as human expert metrics. We introduce detailed settings in Appendix~\ref{geval}.

\subsection{Evaluations}

We used the \textbf{Kendall-Tau correlation} to assess the overall alignment with human judgment, following the approach of previous work by \citet{zuo-etal-2024-patenteval}. This correlation metric evaluates the consistency of the global ranking while disregarding minor errors in individual predictions. We additionally report the \textbf{Spearman correlation}. Compared to Kendall-Tau, Spearman is more sensitive to large rank differences, providing a complementary perspective on the metric ability to predict relative claim quality.

Since the dataset originally presents a three-way classification problem, we also use \textbf{accuracy} and \textbf{F1 scores} to assess model performance. These metrics reflect the model's ability to make precise decisions for individual input pairs, providing a more comprehensive view of its effectiveness. Classification labels can be obtained directly from G-Eval-4, while for other metrics, we assume quality scores to be equivalent if the score differences are less than $10^{-4}$.

\begin{table*}[!ht]
\centering
\footnotesize
\resizebox{.98\textwidth}{!}{
\begin{tabular}{l|l|cc|cc|cc|cc|cc}
\toprule
\multirow{2.7}{*}{\textbf{Type}} & \multirow{2.7}{*}{\textbf{Metric}}    & \multicolumn{2}{c|}{\textbf{Completeness}} & \multicolumn{2}{c|}{\textbf{Clarity}} & \multicolumn{2}{c|}{\textbf{Consistency}} & \multicolumn{2}{c|}{\textbf{Linkage}} & \multicolumn{2}{c}{\textbf{Quality}} \\ \cmidrule{3-12}
& & $\tau$ & $\rho$ & $\tau$ & $\rho$ & $\tau$ & $\rho$ & $\tau$ & $\rho$ & $\tau$ & $\rho$ \\ \midrule
\multirow{6}{*}{N-gram} 
&BLEU-1  & 0.305 & 0.345 & 0.359 & 0.401 & 0.284 & 0.329 & 0.335 & 0.376 & 0.326 & 0.369 \\ 
&BLEU-4  & 0.271 & 0.304 & 0.280 & 0.312 & 0.227 & 0.263 & 0.256 & 0.289 & 0.269 & 0.305 \\ 
&ROUGE-1 & 0.305 & 0.342 & 0.314 & 0.351 & 0.238 & 0.279 & 0.301 & 0.341 & 0.292 & 0.332 \\ 
&ROUGE-2 & 0.305 & 0.342 & 0.280 & 0.312 & 0.215 & 0.251 & 0.268 & 0.303 & 0.269 & 0.306 \\ 
&ROUGE-L & 0.282 & 0.317 & 0.280 & 0.312 & 0.261 & 0.303 & 0.346 & \underline{0.391} & 0.303 & 0.344 \\ 
&METEOR  & 0.316 & 0.358 & 0.371 & 0.414 & \underline{0.307} & \underline{0.355} & 0.324 & 0.364 & 0.292 & 0.331 \\ \midrule
\multirow{4.2}{*}{Embedding} 
&BERTScore & 0.241 & 0.279 & 0.251 & 0.281 & 0.242 & 0.283 & 0.272 & 0.303 & 0.239 & 0.268 \\ 
&BARTScore  & 0.165 & 0.188 & 0.130 & 0.146 & 0.211 & 0.242 & 0.196 & 0.219 & 0.164 & 0.185 \\ 
&MoverScore & 0.199 & 0.227 & 0.199 & 0.217 & 0.223 & 0.264 & 0.231 & 0.265 & 0.210 & 0.243 \\ 
&SimCSE  & 0.177 & 0.196 & 0.165 & 0.173 & 0.143 & 0.165 & 0.220 & 0.246 & 0.165 & 0.186 \\ \midrule
\multirow{2}{*}{Miscellaneous} 
&UniEval  & 0.339 & 0.383 & 0.337 & 0.375 & 0.261 & 0.302 & 0.301 & 0.338 & \underline{0.337} & \underline{0.381} \\
&AlignScore  & 0.146 & 0.162 & 0.145 & 0.160 & 0.261 & 0.305 & 0.200 & 0.226 & 0.224 & 0.255 \\ \midrule
LLM-as-a-judge & G-Eval-4 & \underline{0.377} & \underline{0.410} & \underline{0.412} & \underline{0.481} & 0.276 & 0.353 & \underline{0.350} & 0.385 & 0.277 & 0.310 \\ \midrule
\rowcolor{gray!30}
Ours & PatClaimEval  & \textbf{0.400} & \textbf{0.504} & \textbf{0.461} & \textbf{0.518} & \textbf{0.354} & \textbf{0.424} & \textbf{0.419} & \textbf{0.518} & \textbf{0.477} & \textbf{0.602} \\

\bottomrule
\end{tabular}
}
\caption{Kendall-Tau ($\tau$) and Spearman ($\rho$) correlation of automated metrics with human evaluation results. The highest number in each criterion is in \textbf{bold}, and the second-best result is \underline{underlined}. PatClaimEval shows the highest correlation with human assessments in all criteria.}
\label{table:correlation}
\end{table*}

\section{Results}

\subsection{Correlations with Human Evaluations}

Table \ref{table:correlation} presents the Kendall-Tau and Spearman correlation between different automated metrics and human evaluation results across five criteria: feature completeness, conceptual clarity, terminology consistency, logical linkage, and overall quality. 

Overall, \textbf{PatClaimEval demonstrates the highest correlation with human evaluations across all criteria}, suggesting its effectiveness in evaluating patent claim quality. For feature completeness, PatClaimEval achieves a correlation of $\tau=0.400$ and $\rho=0.504$, which outperforms all other metrics. This finding holds consistently across other criteria, with correlations of $\tau=0.461$ and $\rho=0.518$ for clarity, $\tau=0.354$ and $\rho=0.424$ for consistency, $\tau=0.419$ and $\rho=0.518$ for linkage, and $\tau=0.477$ and $\rho=0.602$ for overall quality. Notably, these values are not only the highest but also significantly surpass existing metrics in their alignment with human judgments. Particularly in overall quality, PatClaimEval outperforms the second-best result by approximately 41.5\% and 58.0\% for Kendall-Tau and Spearman correlation respectively. 

In addition, \textbf{N-gram-based metrics demonstrate relatively higher correlations than embedding-based methods in evaluating patent claims.} While N-gram-based methods can sometimes achieve correlation scores exceeding 0.3 in different evaluation aspects, embedding-based metrics rarely surpass this threshold. For instance, ROUGE-L achieves the second-highest Spearman correlation in logical linkage with a score of 0.391.
N-gram-based methods rely on surface-level overlap between generated and reference text, without capturing semantic information or contextual relevance. These methods typically underperform compared to embedding-based approaches, which calculate semantic similarities, in standard text evaluation tasks \citep{zhang2019bertscore, zhao-etal-2019-moverscore, NEURIPS2021_e4d2b6e6}. However, patent claim evaluation results diverge from these prior findings due to its unique focus on patent examination criteria. Both reference and candidate claims describe the same invention but often use different expressions. In this context, high semantic similarity does not necessarily indicate adherence to patent requirements, resulting in weak correlations with human judgments.
In contrast, gold-standard patent claims use precise language and expressions designed to meet examination standards. Thus, more overlaps with these gold claims may better reflect higher quality, potentially explaining why simple overlap-based methods outperform embedding-based similarity approaches in this domain.
These findings extend to metrics of UniEval and AlignScore. While AlignScore assesses factual consistency and shows less correlation, UniEval that measures relevance between candidate and reference claims performs relatively better. 

\textbf{G-Eval-4 shows strong performance in evaluating completeness, clarity, and linkage.}
G-Eval-4 achieves correlation scores of $\tau=0.377$ and $\rho=0.410$ for completeness, which surpasses all other metrics except for PatClaimEval and is consistent with findings from prior research \citep{jiang-etal-2025-patent}. The high correlation in feature completeness can be attributed to GPT-4's proven capabilities in information extraction \citep{achiam2023gpt, li2023evaluating}. In the context of claim evaluation, GPT-4 effectively extracts key features from both reference and candidate claims. In consequence, it enables accurate comparisons and reaches high scores in feature completeness assessments.
However, its performance in terminology consistency and overall quality is less impressive. A plausible explanation is that GPT-4 is not extensively trained on patent-specific texts, which limits its ability to comprehend the unique linguistic and structural requirements of patent claims. Consequently, relying solely on prompting without further fine-tuning may be insufficient for accurately evaluating patent claims.

\begin{table*}[!ht]
\centering
\footnotesize
\begin{tabular}{l|l|cc|cc|cc|cc|cc}
\toprule
\multirow{2.7}{*}{\textbf{Type}} & \multirow{2.7}{*}{\textbf{Metric}}    & \multicolumn{2}{c|}{\textbf{Completeness}} & \multicolumn{2}{c|}{\textbf{Clarity}} & \multicolumn{2}{c|}{\textbf{Consistency}} & \multicolumn{2}{c|}{\textbf{Linkage}} & \multicolumn{2}{c}{\textbf{Quality}} \\ \cmidrule{3-12}
& & Acc & F1 & Acc & F1 & Acc & F1 & Acc & F1 & Acc & F1  \\ \midrule
\multirow{6}{*}{N-gram} 
&BLEU-1 & 50.5 & 42.6 & 54.3 & 46.7 & 47.8 & 39.2 & 53.8 & 46.5 & 52.2 & 44.3 \\ 
&BLEU-4 & 48.9 & 41.4 & 50.5 & 43.5 & 45.1 & 37.0 & 50.0 & 43.2 & 49.5 & 42.0 \\ 
&ROUGE-1 & 50.5 & 42.8 & 52.2 & 44.8 & 45.7 & 37.5 & 52.2 & 45.0 & 50.5 & 42.9 \\ 
&ROUGE-2 & 50.5 & 42.8 & 50.5 & 43.4 & 44.6 & 36.6 & 50.5 & 43.6 & 49.5 & 42.0 \\ 
&ROUGE-L & 49.5 & 41.8 & 50.5 & 43.5 & 46.7 & 38.3 & \underline{54.3} & 46.9 & 51.1 & 43.4 \\ 
&METEOR & 51.1 & 43.2 & 54.9 & 47.2 & \underline{48.9} & 40.1 & 53.3 & 46.0 & 50.5 & 42.8 \\ \midrule
\multirow{4.2}{*}{Embedding} 
&BERTScore & 46.7 & 39.1 & 48.9 & 42.2 & 45.7 & 37.6 & 51.1 & 44.8 & 48.4 & 41.8 \\
&BARTScore & 43.5 & 35.9 & 42.9 & 36.3 & 44.6 & 36.4 & 47.3 & 40.7 & 44.6 & 37.5 \\
&MoverScore & 45.1 & 38.4 & 46.7 & 41.1 & 44.6 & 36.8 & 48.4 & 42.0 & 46.2 & 39.5 \\
&SimCSE & 44.6 & 38.4 & 45.7 & 40.7 & 41.3 & 34.6 & 48.4 & 42.6 & 44.6 & 38.6 \\ \midrule
\multirow{2}{*}{Miscellaneous} 
&UniEval & 52.2 & 44.1 & 53.3 & 45.8 & 46.7 & 38.3 & 52.2 & 45.0 & \underline{52.7} & 44.8 \\
&AlignScore & 42.9 & 36.4 & 44.0 & 38.0 & 46.7 & 38.3 & 47.3 & 40.8 & 47.3 & 40.1 \\ \midrule 
LLM-as-a-judge 
& G-Eval-4 & \textbf{54.8} & \textbf{54.8} & \underline{55.6} & \underline{55.9} & 45.9 & \underline{43.7} & \textbf{54.8} & \underline{54.6} & 49.6 & \underline{46.9} \\  \midrule
\rowcolor{gray!30}
Ours & PatClaimEval & \underline{52.7} & \underline{53.2} & \textbf{60.3} & \textbf{59.5} & \textbf{50.0} & \textbf{49.3} & 52.7 & \textbf{54.7} & \textbf{56.5} & \textbf{57.4} \\
\bottomrule
\end{tabular}
\caption{Accuracy (Acc \%) and F1 scores (F1 \%) on each evaluation criterion. The highest number in each column is in \textbf{bold}, and the second-best result is \underline{underlined}. PatClaimEval demonstrates relatively high and balanced accuracy and F1 scores across all evaluation criteria.}
\label{table:accuracy}
\end{table*}

\subsection{Classification Performance}

Table \ref{table:accuracy} presents the accuracy and F1 scores of different metrics on each evaluation criterion as a classification problem, including feature completeness, conceptual clarity, terminology consistency, logical linkage, and overall quality. 

\textbf{PatClaimEval achieves the highest accuracy and F1 scores across nearly all evaluation criteria.}
Specifically, for conceptual clarity, PatClaimEval achieves an accuracy of 60.3\% and an F1 score of 59.5\%, outperforming all other metrics. This superior performance extends to consistency and overall quality, where PatClaimEval consistently outperforms other methods. In feature completeness, G-Eval-4 demonstrates slightly better performance to PatClaimEval, with both accuracy and F1 scores of 54.8\%.
Despite these strengths, PatClaimEval's absolute scores in some evaluation criteria, such as consistency, remain modest (50.0\% accuracy and 49.3\% F1 score). The moderate absolute scores indicate potential for improvement, such as expanding dataset sizes,  larger models, or more sophisticated training strategies. Nonetheless, PatClaimEval represents a significant advancement as it achieves a 3.8\% improvement in accuracy and a 10.5\% increase in F1 score over the second-best method for overall quality evaluation. It currently stands as the most effective approach for patent claim evaluation.

\textbf{PatClaimEval and G-Eval-4 exhibit balanced performance between accuracy and F1 scores. }
Both models achieve similar accuracy and F1 scores across all five evaluation criteria, in contrast to other metrics, where F1 scores are normally notably lower than their accuracies. This balance reflects an effective trade-off between precision and recall. Although G-Eval-4 does not lead in accuracy across all aspects, its F1 scores are consistently higher than other metrics except for PatClaimEval.
Based on a careful examination of the results, we attribute this strength to G-Eval-4's ability to handle "equal cases", in which two candidate claims receive identical quality scores. Metrics such as N-gram-based and embedding-based methods struggle to evaluate such cases effectively, resulting in discrepancies between their accuracy and F1 scores. The balanced performance of PatClaimEval and G-Eval-4 highlights their robustness and reliability in patent claim evaluation.

\subsection{Qualitative Analysis}

We show an example of claim comparison in Table~\ref{table:example} to demonstrate the inherent challenges of this task. We identify three types of differences between generated claims B and C. First, Claim C demonstrates higher clarity and language precision. It correctly uses \textit{an annular edge}, whereas Claim B incorrectly uses \textit{a annular edge}, a basic grammatical error. Furthermore, in Claim 3, C uses \textit{further comprising}, which aligns with the gold claim and drafting conventions, while B uses the inappropriate \textit{comprises}. Second, Claim C exhibits a stronger logical linkage between components. It introduces dependent clauses in Claim 3 properly using \textit{wherein} that preserves structural relationships between features, whereas Claim B omits such linkages. Third, Claim C uses the phrase \textit{are configured to} when describing some features. While this phrasing deviates from the gold claim, it does not degrade the quality. Overall, Claim C is better than Claim B. However, current metrics cannot capture such subtle and special differences, which could lead to unreliable performance in claim evaluation.

\section{Conclusion}

We introduce Patent-CE, the first comprehensive benchmark for evaluating patent claims. Patent-CE includes comparative evaluations annotated by patent experts, which focus on five key criteria that align with established patent examination standards: feature completeness, conceptual clarity, terminology consistency, logical linkage, and overall quality. 
Moreover, we propose PatClaimEval, a novel multi-dimensional evaluation method specifically designed for patent claims. Extensive experiments demonstrate the effectiveness of PatClaimEval. It achieves the highest correlation with human expert evaluations across all assessment criteria when compared to existing metrics. This research provides valuable resources for developing automated evaluation methods of patent claims and establishes a solid foundation for more reliable assessments of claim generation systems.

\section*{Limitations}
\label{limitation}
We acknowledge several limitations in this research. Firstly, the dataset used in this study includes only patents documented in English, which may affect the applicability to patents in other languages. Furthermore, our evaluation approach relies on a gold standard. It provides a more reliable way to evaluate patent claims and is especially useful when developing related models for claim generation. However, real-world patent examinations by patent offices consider a range of criteria, including novelty, non-obviousness, and language requirements, without necessarily referencing a predefined gold standard. Patents are evaluated based on their intrinsic merit and their relation to prior art. Therefore, exploring reference-free evaluation approaches for patent claims is an important and worthwhile direction for future work. In addition, better semantic methods or different kinds of CoT prompting strategies may also be worth investigating.

\section*{Ethics Statement}
GPT-4 is under a commercial license provided by OpenAI, and we access it through its API. The use of existing artifacts, including models, evaluation metrics, and datasets, is consistent with their intended use. Our proposed dataset is used for patent claim generation evaluation and will be released under \textit{CC-BY-NC-4.0} license. This dataset does not include potential personal information or offensive content, and no ethics review board was involved. 

\bibliography{anthology,custom}

\appendix

\section{Human Annotations}
\label{humanannotation}

We invite licensed patent attorneys for human evaluations. These professionals are provided with reference claims and candidate claims for assessment. They are informed about the intended use of the evaluation results. Table~\ref{tab:humanrating} outlines the detailed evaluation criteria, aligned with prior research \cite{jiang-etal-2025-large}.  We compare the scores and construct the comparative evaluation dataset. 

\begin{table*}[p]
\centering
\footnotesize
\begin{tabular}{|>{\raggedright\arraybackslash}p{3cm}|>{\raggedright\arraybackslash}p{12cm}|}
\toprule
\textbf{Criteria} & \textbf{Rating Description} \\
\midrule

\textbf{Feature Completeness} &
\begin{itemize}
    \item \textbf{0-2:} Most essential features are missing or poorly described.
    \item \textbf{3-4:} Some essential features are present but significant gaps remain.
    \item \textbf{5-6:} Majority of essential features are covered but with minor omissions.
    \item \textbf{7-8:} Almost all essential features are well described with very few gaps.
    \item \textbf{9-10:} All essential features are thoroughly and comprehensively covered.
\end{itemize} \\
\midrule

\textbf{Conceptual Clarity} &
\begin{itemize}
    \item \textbf{0-2:} Claims are very unclear and ambiguous.
    \item \textbf{3-4:} Claims have significant clarity issues, making them difficult to understand.
    \item \textbf{5-6:} Claims are mostly clear but contain some ambiguous language.
    \item \textbf{7-8:} Claims are clear with minimal ambiguity.
    \item \textbf{9-10:} Claims are exceptionally clear and completely unambiguous.
\end{itemize} \\
\midrule

\textbf{Terminology Consistency} &
\begin{itemize}
    \item \textbf{0-2:} Terminology is highly inconsistent.
    \item \textbf{3-4:} Significant inconsistencies in terminology.
    \item \textbf{5-6:} Some inconsistencies in terminology but mostly uniform.
    \item \textbf{7-8:} Terminology is largely consistent with minor inconsistencies.
    \item \textbf{9-10:} Terminology is completely consistent throughout.
\end{itemize} \\
\midrule

\textbf{Logical Linkages} &
\begin{itemize}
    \item \textbf{0-2:} Features are poorly linked with many inaccuracies.
    \item \textbf{3-4:} Significant issues with the linkages of features.
    \item \textbf{5-6:} Mostly accurate linkages with some incorrect connections.
    \item \textbf{7-8:} Accurate linkages with minor inaccuracies.
    \item \textbf{9-10:} Features are accurately and correctly linked throughout.
\end{itemize} \\
\midrule

\textbf{Overall Quality} &
\begin{itemize}
    \item Calculated by: $(completeness*4 + clarity*2 + consistency*2 + correctness*3) \div 11$
    \item \textbf{0-2:} Very poor overall quality, fails to meet most criteria.
    \item \textbf{3-4:} Low overall quality with significant issues across multiple criteria.
    \item \textbf{5-6:} Average overall quality, meets criteria at a basic level.
    \item \textbf{7-8:} High overall quality with minor issues.
    \item \textbf{9-10:} Excellent overall quality, meets or exceeds all criteria.
\end{itemize} \\
\bottomrule

\end{tabular}
\caption{Rating criteria for human annotation deriving from \citet{jiang-etal-2025-large}}
\label{tab:humanrating}
\end{table*}

\section{Dataset Statistics}
\label{moredatastats}

We report the token length statistics of the Patent-CE dataset using the Longformer tokenizer. The results are summarized as follows: the minimum length is 156 tokens, the maximum length is 1,461 tokens, the average length is 644 tokens, the median length is 631 tokens, and the standard deviation is 245 tokens. All claims fall within the token limit of Longformer (4096 tokens), and thus no truncation or segmentation strategies were used. This ensures that input length limitations do not affect the evaluation results. Since the dataset does not include very long claims, the proposed method may not generalize well to extremely long claims that exceed the model’s input capacity.

\section{Experimental Details}
\label{experimentdetails}

All training and testing processes are conducted on NVIDIA A100 GPUs. The total running time is about 20 hours. We randomly select 10\% samples from the training set as the validation set. During training, we use a batch size of 4, a learning rate of 5e-6, a weight decay of 0.01, and training epochs of 10. 

For BLEU, ROUGE, METEOR, and BERTScore, we use the package from the HuggingFace \textit{evaluate} library.\footnote{\url{https://github.com/huggingface/evaluate}} For MoverScore\footnote{\url{https://github.com/AIPHES/emnlp19-moverscore}}, BARTScore\footnote{\url{https://github.com/neulab/BARTScore}}, AlignScore\footnote{\url{https://github.com/yuh-zha/AlignScore}}, SimCSE\footnote{\url{https://github.com/princeton-nlp/SimCSE}}, and UniEval\footnote{\url{https://github.com/maszhongming/UniEval}}, we use their code from original repositories. 
We use the \textit{scipy} Python library to calculate the correlation scores and  \textit{scikit-learn} for accuracy and F1 scores. 

\section{G-Eval-4}
\label{geval}

\begin{table*}[!t]
\centering
\footnotesize
\begin{tabular}{|p{0.95\linewidth}|}
\toprule
\textbf{Instructions:} \\
You will be given the draft claims and the referenced claims of the same patent. Your task is to rate the draft claims on four metrics using the referenced claims as the gold standard. Please make sure you read and understand these instructions carefully. Keep this document open while reviewing, and refer to it as needed. \\
\\
\textbf{Evaluation Criteria:} \\
\textbf{1. Completeness of Essential Features (0--100)}\\
The extent to which the generated claims encapsulate all critical aspects of the invention. \\
\quad 0--20: Most essential features are missing or poorly described. \\
\quad 21--40: Some essential features are present but significant gaps remain. \\
\quad 41--60: Majority of essential features are covered but with minor omissions. \\
\quad 61--80: Almost all essential features are well described with very few gaps. \\
\quad 81--100: All essential features are thoroughly and comprehensively covered. \\
\\
\textbf{2. Conceptual Clarity (0--100)} \\
The clarity and unambiguity of the language used in the claims. \\
\quad 0--20: Claims are very unclear and ambiguous. \\
\quad 21--40: Claims have significant clarity issues, making them difficult to understand. \\
\quad 41--60: Claims are mostly clear but contain some ambiguous language. \\
\quad 61--80: Claims are clear with minimal ambiguity. \\
\quad 81--100: Claims are exceptionally clear and completely unambiguous. \\
\\
\textbf{3. Consistency in Terminology (0--100)}\\
The uniformity in the use of terms throughout the claims. \\
\quad 0--20: Terminology is highly inconsistent. \\
\quad 21--40: Significant inconsistencies in terminology. \\
\quad 41--60: Some inconsistencies in terminology but mostly uniform. \\
\quad 61--80: Terminology is largely consistent with minor inconsistencies. \\
\quad 81--100: Terminology is completely consistent throughout. \\
\\
\textbf{4. Technical Correctness of Feature Linkages (0--100)} \\
The accuracy with which the features are interconnected and related. \\
\quad 0--20: Features are poorly linked with many inaccuracies. \\
\quad 21--40: Significant issues with the linkages of features. \\
\quad 41--60: Mostly accurate linkages with some incorrect connections. \\
\quad 61--80: Accurate linkages with minor inaccuracies. \\
\quad 81--100: Features are accurately and correctly linked throughout. \\
\\
\textbf{Evaluation Steps:} \\
1. Read the referenced claims carefully and identify the invention's features. Assume the referenced claims have scores of 100 in all Evaluation Criteria. \\
2. Read the draft claims and compare them to the referenced claims. \\
3. Assign a score for each metric based on the Evaluation Criteria. \\
\\
\textbf{Example:} \\
Referenced Claims: \texttt{<<Claims>>} \\
Draft Claims: \texttt{<<Claims>>} \\
Evaluation Form (scores ONLY): \\
- Completeness of Essential Features: X \\
- Conceptual Clarity: X \\
- Consistency in Terminology: X \\
- Technical Correctness of Feature Linkages: X \\
\bottomrule
\end{tabular}

\caption{G-Eval prompt used for claim evaluation originated from \citet{jiang-etal-2025-patent}}
\label{tab:geval}

\end{table*}

We use the following prompt for G-Eval consistent with previous research \citep{jiang-etal-2025-patent}, as shown in Table~\ref{tab:geval}.
We use GPT-4 to evaluate feature completeness, conceptual clarity, terminology consistency, and logical linkage. The overall quality is calculated based on the same formula of human evaluation in Table \ref{tab:humanrating}. 

\section{Example Claim Comparison}

We present an example of claim comparison in Table~\ref{table:example}, where the differences between Claim B and C are marked in blue. 

\begin{table*}[!ht]
\centering
\footnotesize
\resizebox{\textwidth}{!}{

\begin{tabular}{|p{15.5cm}|}
\toprule

\textbf{Gold Claim A} \\
1. A shroud for connecting to a container having a closure portion, the shroud comprising: a housing having a luer connector; a spike having a fluid lumen transitioning into the connector; a plurality of segments terminating in \textcolor{blue}{a continuous annular edge} surrounding the spike and defining a plurality of openings; and a plurality of protrusions circumferentially spaced, each of the plurality of protrusions having (i) a proximal end connected to the continuous annular edge and (ii) a distal end positioned in one of the plurality of openings, the distal end of each protrusion inwardly extending from the continuous annular edge, the protrusions defining a receiving volume dimensioned to receive the closure portion of the container, whereby the \textcolor{blue}{protrusions deflect} outwardly away from an initial position during contact and traversing of the closure portion of the container and return to approximately their initial position when the closure portion of the container passes the distal end of the protrusion, whereby the \textcolor{blue}{protrusions deflect} inwardly towards the closure portion of the container when a removing force is applied to the container, thereby preventing removal of the container. \\
2. The shroud of claim 1, wherein the housing comprises a surface with parallel raised features for facilitating gripping by a user. \\
3. The shroud of claim 1, further \textcolor{blue}{comprising} a vent housing and \textcolor{blue}{wherein} the spike further comprises a vent lumen connected to the vent housing. \\
4. The shroud of claim 3, wherein the vent housing further comprises a filter. \\ 

\midrule

\textbf{Generated Claim B} \\
A shroud for connecting to a container having a closure portion, the shroud comprising: a housing having a connector; a spike having a fluid lumen fluidically coupled to the connector; a plurality of segments terminating in \textcolor{blue}{a annular edge} surrounding the spike and defining a plurality of openings; a plurality of protrusions circumferentially spaced and inwardly extending from the annular edge, the protrusions defining a receiving volume dimensioned to receive the closure portion of the container, wherein the protrusions have a proximal end connected to the annular edge and a distal end positioned in the opening, whereby the \textcolor{blue}{protrusions deflect} outwardly away from an initial position during contact and traversing of the closure portion of the container and return to approximately their initial position when the closure portion of the container passes the distal end of the protrusion, whereby the \textcolor{blue}{protrusions deflect} inwardly towards the closure portion of the container when a removing force is applied to the container, thereby preventing removal of the container. \\
2. The shroud of claim 1, wherein the housing comprises a surface with raised features for facilitating gripping by a user. \\
3. The shroud of claim 1 further \textcolor{blue}{comprises} a vent housing and the spike further comprises a vent lumen coupled to the vent housing. \\
4. The shroud of claim 3, wherein the vent housing further comprises a filter. \\
\midrule

\textbf{Generated Claim C} \\

1. A shroud for connecting to a container having a closure portion, the shroud comprising: a housing having a connector; a spike having a fluid lumen fluidically coupled to the connector; a plurality of segments terminating in \textcolor{blue}{an annular edge} surrounding the spike and defining a plurality of openings; a plurality of protrusions circumferentially spaced and inwardly extending from the annular edge, the protrusions defining a receiving volume dimensioned to receive the closure portion of the container, wherein the protrusions have a proximal end connected to the annular edge and a distal end positioned in the opening, whereby the \textcolor{blue}{protrusions are configured to deflect} outwardly away from an initial position during contact and traversing of the closure portion of the container and return to approximately their initial position when the closure portion of the container passes the distal end of the protrusion, whereby the \textcolor{blue}{protrusions are further configured to deflect} inwardly towards the closure portion of the container when a removing force is applied to the container, thereby preventing removal of the container. \\
2. The shroud of claim 1, wherein the housing comprises a surface with raised features for facilitating gripping by a user. \\
3. The shroud of claim 1, further \textcolor{blue}{comprising} a vent housing and \textcolor{blue}{wherein} the spike further comprises a vent lumen coupled to the vent housing. \\
4. The shroud of claim 3, wherein the vent housing further comprises a filter.
  \\

\bottomrule
\end{tabular}
}
\caption{An example of claim comparison. Differences between B and C are marked in \textcolor{blue}{blue}, and C is better.}
\label{table:example}
\end{table*}

\end{document}